\documentclass{article}
\pdfoutput=1
\usepackage{ijcai25}

\usepackage{times}
\usepackage{soul}
\usepackage{url}
\usepackage[hidelinks]{hyperref}
\usepackage[utf8]{inputenc}
\usepackage[small]{caption}
\usepackage{graphicx}
\usepackage{amsmath}
\usepackage{amsthm}
\usepackage{booktabs}
\usepackage{algorithm}
\usepackage{algorithmic}
\usepackage{mydef}
\usepackage[switch]{lineno}

\title{A Survey of Large Language Model-Powered Spatial Intelligence Across Scales: Advances in Embodied Agents, Smart Cities, and Earth Science}

\author{
    Jie Feng, Jinwei Zeng, Qingyue Long, $^\beta$Hongyi Chen, Jie Zhao, $^\dagger$Yanxin Xi, Zhilun Zhou, \\ Yuan Yuan, $^\S$Shengyuan Wang, Qingbin Zeng, Songwei Li, Yunke Zhang, \\
    Yuming Lin, Tong Li, Jingtao Ding, Chen Gao, Fengli Xu, Yong Li \\
    \affiliations
    Department of Electronic Engineering, BNRist, Tsinghua University, Beijing, China, \\ 
  $^\beta$ Shenzhen International Graduate School, Tsinghua University, Shenzhen, China,  \\ %
  $^\S$Department of Computer Science, Tsinghua University, Beijing, China, \\
  $^\dagger$Department of Computer Science, University of Helsinki, Helsinki, Finland  \\
    \emails
    \{fengjie, liyong07\}@tsinghua.edu.cn
}

\begin{document}

\maketitle

\begin{abstract}
Over the past year, the development of large language models (LLMs) has brought spatial intelligence into focus, with much attention on vision-based embodied intelligence. However, spatial intelligence spans a broader range of disciplines and scales, from navigation and urban planning to remote sensing and earth science. What are the differences and connections between spatial intelligence across these fields? In this paper, we first review human spatial cognition and its implications for spatial intelligence in LLMs. We then examine spatial memory, knowledge representations, and abstract reasoning in LLMs, highlighting their roles and connections. Finally, we analyze spatial intelligence across scales—from embodied to urban and global levels—following a framework that progresses from spatial memory and understanding to spatial reasoning and intelligence. Through this survey, we aim to provide insights into interdisciplinary spatial intelligence research and inspire future studies.
\end{abstract}

\renewcommand\thefootnote{} 
\footnotetext{All authors contribute equally to this work.}

\section{Introduction}

Spatial intelligence is an inherently interdisciplinary research field, encompassing diverse challenges, application scenarios, and methodologies across multiple domains. For example, navigating within a room requires spatial intelligence, designing a 15-minute community relies on spatial intelligence, predicting the possible location of an image involves spatial intelligence, and analyzing the spatial patterns of climate is also a form of spatial intelligence. In other words, spatial intelligence is ubiquitous and plays a crucial role in human society and physical world.

\begin{figure*}[ht]
   \centering
    \includegraphics[width=0.8\linewidth]{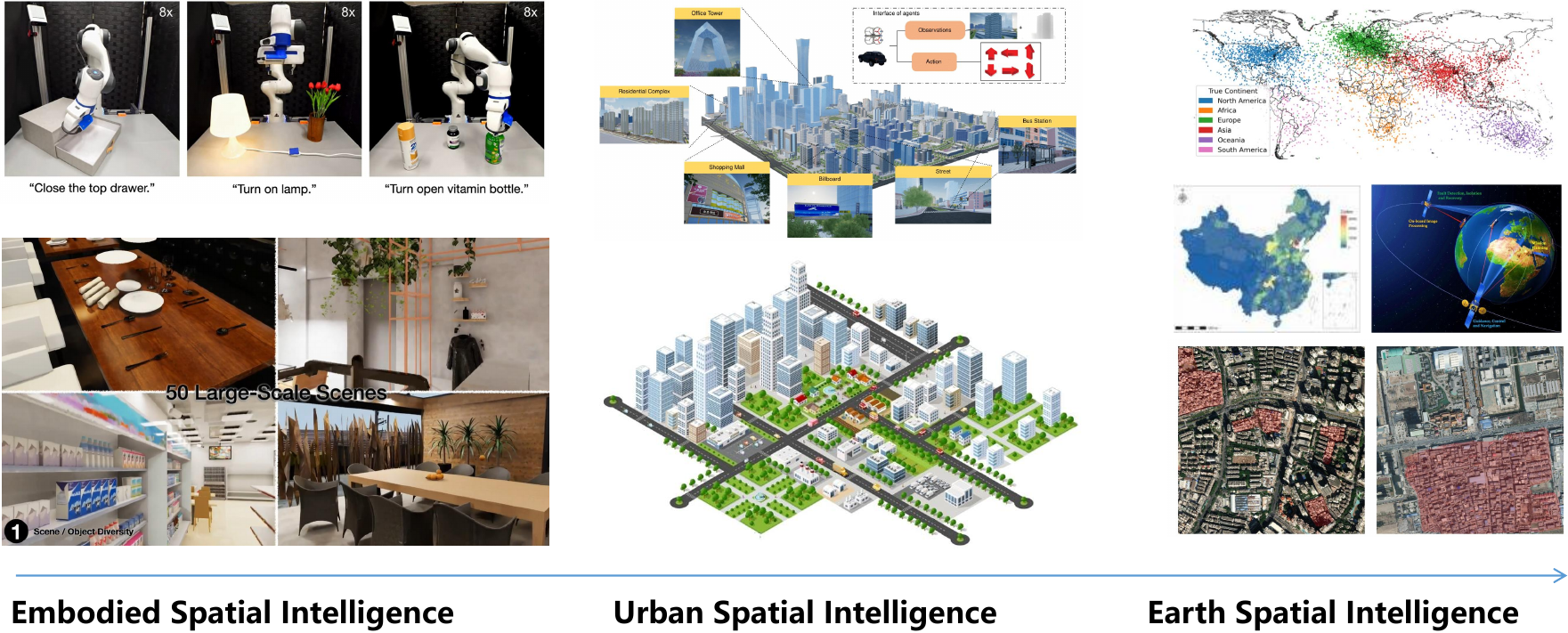}
    \caption{Multiple scale spatial intelligence in real world: from embodied spatial intelligence to earth spatial intelligence.}
    \label{fig:framework}
\end{figure*}

Research on spatial intelligence has deep historical roots. On the one hand, it serves as a crucial avenue for humans to understand their own cognitive and perceptual mechanisms~\cite{ishikawa2021spatial,eichenbaum2014can}. Studies on human spatial cognition, ranging from mental mapping to wayfinding strategies, have provided foundational insights into human intelligence. On the other hand, spatial intelligence has long had practical significance in real-world applications, such as embodied navigation~\cite{lin2024advancese}, geographic information systems (GIS)~\cite{zhao2024artificial}, and climate prediction~\cite{she2024llmdiff}. The study of spatial intelligence continues to evolve, bridging cognitive science, artificial intelligence, and applied domains.

The rapid advancements in deep learning, particularly in large language models (LLMs), have significantly contributed to spatial intelligence research of recent years. LLMs have made notable progress with world knowledge, planning and reasoning capabilities, and powerful generalization across tasks. These advancements have fueled research in embodied intelligence~\cite{gupta2021embodied}, where LLMs play a central role in areas such as robotic navigation, multimodal perception, and control. Recent works, such as SpatialVLM~\cite{chen2024spatialvlm} and Voxposer~\cite{huang2023voxposer}, have demonstrated how LLMs can improve spatial reasoning and decision-making in embodied agents, enabling them to operate more effectively in complex environments.

Beyond embodied intelligence, LLMs have also inspired new research in urban and global-scale spatial intelligence. In urban research, for example, LLMs have been integrated with geospatial data to optimize urban planning~\cite{zhou2024large}, traffic prediction~\cite{li2024urbangpt} and infrastructure management~\cite{lai2023large}. At a global scale, researchers have explored how LLMs can enhance remote sensing analysis~\cite{kuckreja2024geochat} and disaster prediction~\cite{zhang2023skilful}, and so on, which illustrate the potential of LLMs to process large-scale geospatial information and generate meaningful insights for global-scale decision-making. These interdisciplinary applications highlight the transformative impact of LLMs on spatial intelligence research, paving the way for future developments across multiple domains.

Despite the growing of research on spatial intelligence across various fields, there is still a lack of a unified framework for comprehensively understanding and analyzing it. Existing studies often focus on specific aspects, such as vision-based embodied intelligence, urban planning, or remote sensing intelligence, without integrating insights across disciplines and scales. To bridge this gap, this survey traces the development of spatial intelligence from the perspective of human cognition, fundamental spatial capabilities, and multi-scale system intelligence from embodied agents, urban intelligence and earth science. By synthesizing these perspectives, we aim to provide a cohesive foundation for interdisciplinary research, offering insights and inspiration for future advancements in spatial intelligence.

Our survey makes three key contributions. First, it establishes a structured analytical framework for understanding spatial intelligence across diverse disciplines and scales, advancing from spatial memory and perception to reasoning and higher-level intelligence. Second, it synthesizes existing literature on spatial intelligence applications with LLMs across multiple fields, alongside discussions on spatial memory, knowledge representation, and spatial reasoning in LLMs, providing researchers with a timely and valuable reference. Third, it explores key challenges and open questions in interdisciplinary spatial intelligence research, uncovering connections between embodied, urban, and global-scale intelligence while outlining promising directions for future exploration.

\section{Background and Taxonomy}
\tikzstyle{leaf}=[draw=hiddendraw,
    rounded corners,minimum height=1em,
    fill=mygreen!40,text opacity=1, align=center,
    fill opacity=.5,  text=black,align=left,font=\scriptsize,
    inner xsep=3pt,
    inner ysep=1pt,
    ]
\tikzstyle{middle}=[draw=hiddendraw,
    rounded corners,minimum height=1em,
    fill=output-white!40,text opacity=1, align=center,
    fill opacity=.5,  text=black,align=left,font=\scriptsize,
    inner xsep=3pt,
    inner ysep=1pt,
    ]
\begin{figure*}[ht]
\centering
\begin{forest}
  for tree={
  forked edges,
  grow=east,
  reversed=true,
  anchor=base west,
  parent anchor=east,
  child anchor=west,
  base=middle,
  font=\scriptsize,
  rectangle,
  line width=0.7pt,
  draw=output-black,
  rounded corners,align=left,
  minimum width=2em,
    s sep=5pt,
    inner xsep=3pt,
    inner ysep=1pt,
  },
  where level=1{text width=4.5em}{},
  where level=2{text width=5.4em,font=\scriptsize}{},
  where level=3{font=\scriptsize}{},
  where level=4{font=\scriptsize}{},
  where level=5{font=\scriptsize}{},
  [Large Language Model-Empowered Spatial Intelligence, middle,rotate=90,anchor=north,edge=output-black
    [Foundational \\ Capabilities, middle, edge=output-black,text width=5.9em
        [Spatial Memory \\ and Knowledge, middle, text width=5.4em, edge=output-black
            [Internal Encoded~\cite{petroni2019language}{,}~\cite{gurnee2024language}{,}~\cite{roberts2020much}, leaf, text width=31.0em, edge=output-black]
            [Externally Integrated~\cite{mansourian2024chatgeoai}~\cite{yu2024rag}, leaf, text width=31.0em, edge=output-black]
        ]
        [Abstract Spatial \\ Reasoning, middle, text width=5.4em, edge=output-black
            [Qualitative Reasoning~\cite{yamada2023evaluating}{,}~\cite{sharma2023exploring}{,}~\cite{lehnert2024beyond}{,}~\cite{li2024advancing}, leaf, text width=31.0em, edge=output-black]
            [Geometric Reasoning: GeoEval~\cite{zhang2024geoeval}{,} GeomVerse~\cite{kazemi2023geomverse}, leaf, text width=31.0em, edge=output-black],
            [Graph-theoretical Reasoning: GraphInstruct~\cite{luo2024graphinstruct}, leaf, text width=31.0em, edge=output-black]
        ]
    ]
     [Spatial Intelligence \\ for Real World, middle, edge=output-black, text width=5.9em
        [Embodied Spatial \\ Intelligence, middle, text width=5.4em, edge=output-black
            [Spatial Perception and Understanding: LLMI3D~\cite{yang2024llmi3d}{,} 3D-MEM~\cite{yang20243dmem}{,} \\ EmbodiedScan~\cite{wang2024embodiedscan}{,} Scene-LLM~\cite{fu2024scene}{,} SpatialBot~\cite{cai2024spatialbot}, leaf, text width=31.0em, edge=output-black]
            [Spatial Interaction and Navigation: RT-2~\cite{zitkovich2023rt}{,} VIMA~\cite{jiang2022vima}{,}\\
            Guide-LLM~\cite{song2024guide}{,} NavGPT~\cite{zhou2024navgpt}{,} TopV-Nav~\cite{zhong2024topv},
            leaf, text width=31.0em, edge=output-black]
        ],
        [Urban Spatial \\ Intelligence, middle, text width=5.4em, edge=output-black
            [Spatial Understanding and Memory: GeoLLM~\cite{manvi2023geollm}{,} GeoChat~\cite{kuckreja2024geochat}{,}\\  UrbanCLIP~\cite{yan2024urbanclip}{,} ReFound~\cite{xiao2024refound}{,} UrbanKGEnt~\cite{ning2024urbankgent}, leaf, text width=31.0em, edge=output-black]
            [Spatial Reasoning and Intelligence: GeoReasoner~\cite{ligeoreasoner}{,} LLMob~\cite{wang2024large}{,} \\ AgentMove~\cite{feng2024agentmove}{,}  Mobility-LLM~\cite{gong2024mobility}{,} FLAME~\cite{xu2024flame}{,} 
            , leaf, text width=31.0em, edge=output-black]
        ],
        [Earth Spatial \\ Intelligence, middle, text width=5.4em, edge=output-black
            [Global Encoding: TorchSpatial Benchmark~\cite{wu2024torchspatial}, leaf, text width=31.0em, edge=output-black]
            [Climate: LLMDiff~\cite{she2024llmdiff}{,}  CLLMate~\cite{li2024cllmate}{,} GenCast~\cite{ravuri2021skilful}, leaf, text width=31.0em, edge=output-black]
            [Geography: 
            GeoGPT~\cite{zhang2023geogpt}{,} GeoSEE~\cite{han2024geosee}{,} GeoReasoner~\cite{yan2024georeasoner}, leaf, text width=31.0em, edge=output-black]
            [Other Disciplines: OceanPlan~\cite{yang2024oceanplan}{,} Orca~\cite{li2024ocean}{,} MineAgent\cite{yu2024mineagent}, leaf, text width=31.0em, edge=output-black]
        ]
    ]
  ]
\end{forest}
\caption{A taxonomy of large language model-empowered spatial intelligence with representative examples.}
\label{fig:taxonomy_of_pGMs}
\end{figure*}
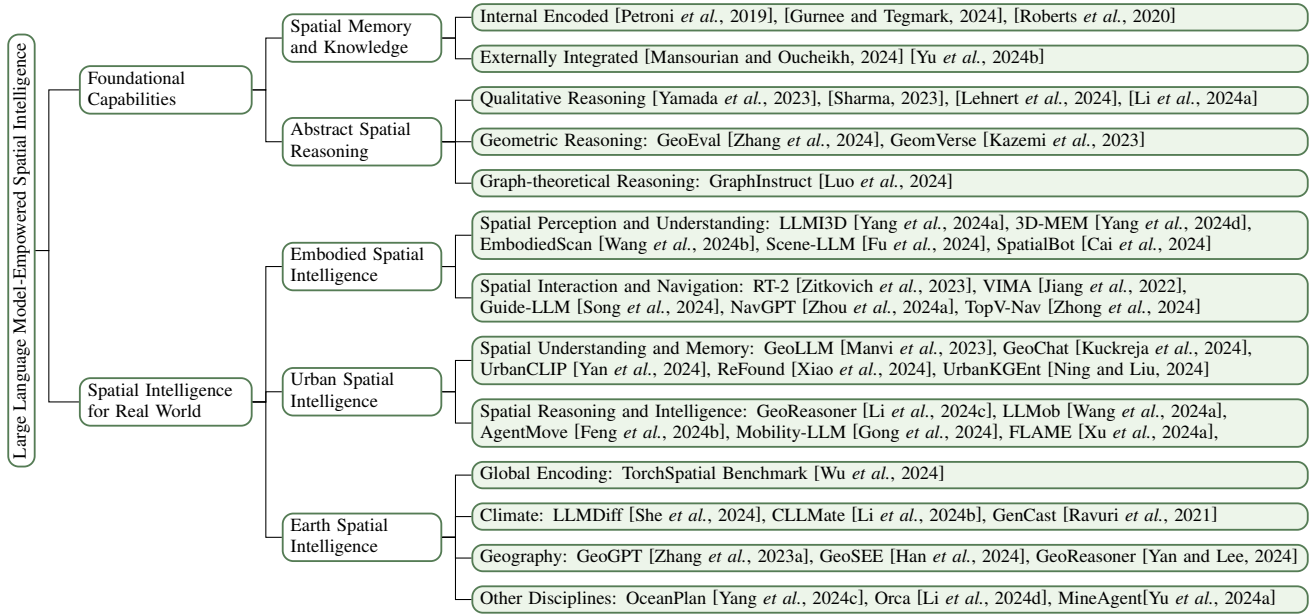

\subsection{Spatial Intelligence of Human} 
Here, we first review human spatial intelligence research from the perspectives of neuroscience and cognitive science, elucidating the potential abilities and origins of spatial intelligence across various domains and scales. Furthermore, we explore the relationship between spatial intelligence and other human intelligences. These findings will enhance our understanding of the critical capabilities of cross-domain spatial intelligence and facilitate the development of more effective methods for constructing and enhancing spatial intelligence.

\subsubsection{Cognitive Map} 
Spatial cognitive map is the internal representation of environmental knowledge, characterized by subjectivity and distortion~\cite{ishikawa2021spatial}. Tolman introduced this concept in 1948~\cite{tolman1948cognitive}, later expanded by Eichenbaum et al.~\cite{cohen1993memory,eichenbaum2014can}, emphasizing the hippocampus's role in spatial and non-spatial memory. 
At the neural level, spatial representation relies on place cells in the hippocampus and grid cells in the entorhinal cortex~\cite{moser2008place,moser2017spatial}. Place cells activate when an individual is in a specific location, while grid cells provide a coordinate-like system for mapping the environment. These cells, along with head direction cells and boundary cells, form the neural basis for constructing spatial cognitive maps~\cite{long2025allocentric}. 
Recent advancements, such as the Tolman-Eichenbaum Machine (TEM)~\cite{whittington2020tolman}, highlight the ability to generalize spatial and relational memory through structural abstraction and cross-environment representation by grid cells. Comparatively, large language models (LLMs) leverage Transformer architectures to emulate spatial tasks, such as positional encoding and navigation, drawing parallels to hippocampal functions~\cite{whittington2021relating}. 

\subsubsection{Spatial Schema}

Schemas are high-level knowledge structures that encapsulate the common features abstracted from multiple experiences. These structures play a critical role in the processes of perceiving, interpreting, and remembering events. They continuously evolve with the accumulation of new experiences and memories, influencing the formation, consolidation, and retrieval of memory~\cite{gilboa2017neurobiology}.
In human spatial cognition, schemas play a crucial role. Spatial schemas are high-level spatial cognitive structures formed through the transfer and generalization of experiences across different environments. Unlike cognitive maps, their processing is centered in specific regions of the neocortex. Spatial schemas are highly abstract in nature, emerging through the integration of overlapping neural representations in similar environments. They serve as higher-order spatial representations that transcend specific environments, such as the anticipated layout of a modern city~\cite{farzanfar2023cognitive}. Spatial schemas and cognitive maps, as distinct levels of spatial cognitive structures, interact and influence each other, jointly contributing to human spatial cognition.

Recent research has explored the similarities and connections between spatial intelligence based on LLMs and human spatial intelligence, e.g., Momennejad et al.~\cite{momennejad2024evaluating} assessed their cognitive mapping capabilities.  However, LLMs exhibit limitations, including topological reasoning errors (e.g., fictitious paths, inefficiency) and visual-spatial perception gaps. While studying cognitive maps in both humans and LLMs provides valuable insights into spatial intelligence, significant challenges remain in enhancing LLMs' schema learning and spatial syntax integration.

\subsection{Taxonomy of Spatial Intelligence}

Building on human spatial memory and intelligence, we propose a taxonomy for spatial memory and intelligence in LLMs, as illustrated in Figure~\ref{fig:taxonomy_of_pGMs}, and provide a comprehensive survey of current research based on this framework. Specifically, we first introduce the foundational capabilities that enable spatial intelligence in LLMs, which are divided into spatial memory and knowledge, as well as abstract spatial reasoning abilities. Subsequently, we focus on the application of spatial intelligence in the real world, exploring three dimensions: embodied intelligence, urban intelligence, and earth intelligence.

\section{Foundational Capabilities of Spatial Intelligence in LLMs}
\begin{figure*}
    \centering
    \includegraphics[width=0.7\textwidth]{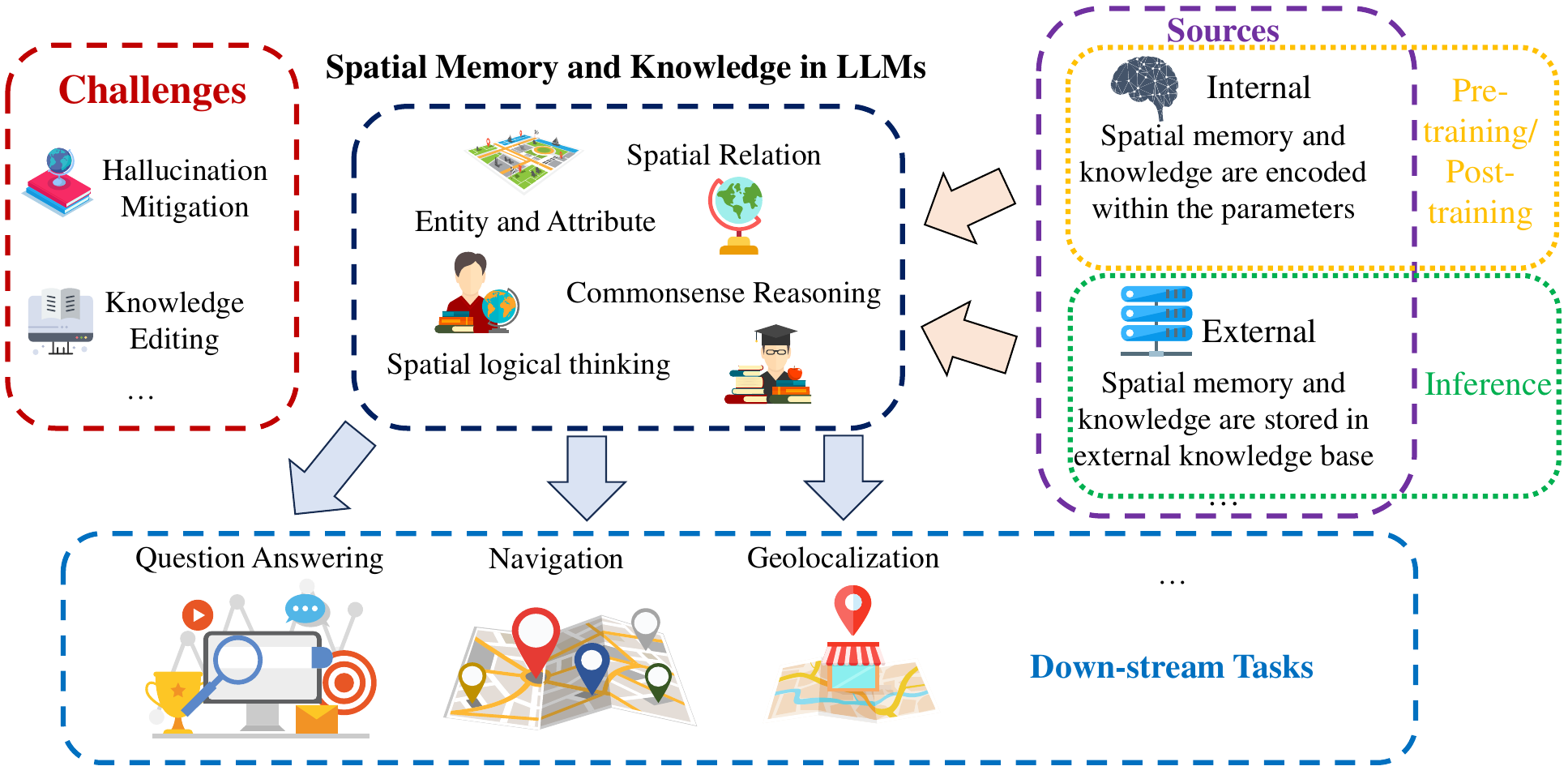}
    \caption{This figure illustrates the core concepts of Spatial Memory and Knowledge in LLMs. LLMs build their spatial memory and knowledge from both internal and external sources to perform tasks like question answering, navigation, and geolocalization, while also facing challenges such as hallucination mitigation and knowledge editing.}
    \label{fig:memory_n_knowledge}
\end{figure*}

\subsection{Spatial Memory and Knowledge in LLMs} 

Spatial memory refers to the cognitive ability to recall spatial relationships, entities, and attributes encountered in the past. Spatial knowledge, a broader concept, encompasses not only this memory but also commonsense reasoning and logical thinking related to space. General spatial memory and knowledge combine both abstract spatial cognition and real-world environmental capabilities.

Recently, state-of-the-art large language models (LLMs) have demonstrated their proficiency in handling spatial tasks with spatial memory and knowledge \cite{bhandari2023large}. Multi-modal large language models (MLLMs) also extend this capability,  exhibiting their memory and knowledge about spatial information from both linguistic and visual modalities \cite{yang2024thinking}. 
Spatial memory and knowledge can be derived from internal or external sources. Internally, spatial memory and knowledge are encoded within the parameters of LLMs during pretraining or post-training stages \cite{petroni2019language,gurnee2024language,roberts2020much}. Externally, LLM's can utilize outer spatial memory or knowledge storage for specific information when needed \cite{mansourian2024chatgeoai}. 
LLMs' spatial memory and knowledge is an essential part of their spatial intelligence. Many general and spatial-specific tasks are based on accurate and adequate memory and knowledge about the spatial environment, including question answering \cite{mai2021geographic,yamada2023evaluating}, navigation\cite{epstein2017cognitive,feng2024citybench}, and geolocalization\cite{haas2024pigeon}. 
Practices to improve LLM's spatial memory and knowledge emerge along the bloom of pre-trained generative models. Various training methods are implemented to encode spatial information \cite{feng2024citygpt}. Other works integrate external knowledge base to provide spatial memory and knowledge \cite{yu2024rag}. Previous works have also attempted to leverage compressed spatial knowledge within LLMs \cite{manvi2023geollm}.

Despite these rapid advancements, challenges remain in the domain of spatial memory and knowledge in LLMs. One significant challenge is hallucination \cite{lee2022factuality}, where LLMs may generate non-factual or non-faithful contents \cite{huang2023survey}, undermining the effectiveness of task in spatial contexts. Another pressing challenge is knowledge editing \cite{zhang2023large}. Given the dynamic nature of the spatial environment, it is necessary to continually and timely update LLM's memory and knowledge to reflect accurate spatial information.

\subsection{Abstract Spatial Reasoning of LLMs} 
Abstract reasoning ability is a crucial cognitive capability that enables intelligent agents to simplify complex reality into operable mental models. In the context of spatial intelligence, abstract reasoning plays a crucial role: it not only simplifies complex physical spaces into manageable mental models but also provides a foundation for higher-level spatial cognition, serving as a vital bridge between objective spatial environments and cognitive representations.

With LLMs showing promise in cognitive tasks, assessing their spatial abstract reasoning capabilities has emerged as a critical research direction, both for understanding their limitations and guiding future improvements.
Current assessments of LLMs' spatial abstract reasoning capabilities primarily focus on three directions: qualitative spatial reasoning~\cite{yamada2023evaluating,sharma2023exploring,lehnert2024beyond,li2024advancing}, geometric reasoning~\cite{zhang2024geoeval,kazemi2023geomverse}, and graph-theoretical reasoning~\cite{luo2024graphinstruct}. Qualitative spatial reasoning evaluates models' ability to understand and reason about spatial relations and transformations through linguistic descriptions. In this domain, LLMs have revealed significant performance degradation in multi-hop reasoning tasks while demonstrating that structured thinking frameworks can effectively mitigate these limitations~\cite{li2024advancing}. In spatial planning problems, ~\cite{lehnert2024beyond} show that training strategies like search dynamics bootstrapping have shown notable improvements in complex spatial planning tasks.
Geometric reasoning focuses on assessing models' understanding of mathematical-geometric concepts and their applications in spatial problem-solving. GeoEval~\cite{zhang2024geoeval} comprehensively evaluates LLMs across various geometry domains and identified their weakness in inverse reasoning compared to forward reasoning while showing the effectiveness of problem rephrasing strategies. GeomVerse~\cite{kazemi2023geomverse} systematically demonstrates VLMs' struggles with deep geometric reasoning tasks requiring long inference chains rather than simple knowledge retrieval.
Graph-theoretical reasoning examines models' capabilities in understanding and manipulating graph structures. In this field, GraphInstruct~\cite{luo2024graphinstruct} developed a comprehensive test set, which revealed that LLMs still struggle with complex graph algorithms like minimum spanning trees, Hamiltonian paths, and shortest paths. However, their research also demonstrated that these limitations can be overcome through structured training approaches that emphasize intermediate reasoning steps. Besides, Xu~\cite{xu2025defining} et al. pioneer a psychometric framework that defines five basic spatial abilities (BSAs) in vision-language models (VLMs), while highlighting issues such as weak geometry encoding and the absence of dynamic simulation capabilities.

In summary, current evaluations across these three directions reveal that pre-trained LLMs primarily rely on language understanding to process abstract spatial problems, lacking genuine spatial cognitive abilities. Methodological improvements, including structured reasoning frameworks, knowledge-guided training, and intermediate process supervision, have shown promise in addressing these limitations. Moving forward, the field requires both more comprehensive evaluation standards and meaningful comparisons with human performance to better understand and advance LLMs' spatial reasoning capabilities.

\begin{figure*}
    \centering
    \includegraphics[width=0.6\linewidth]{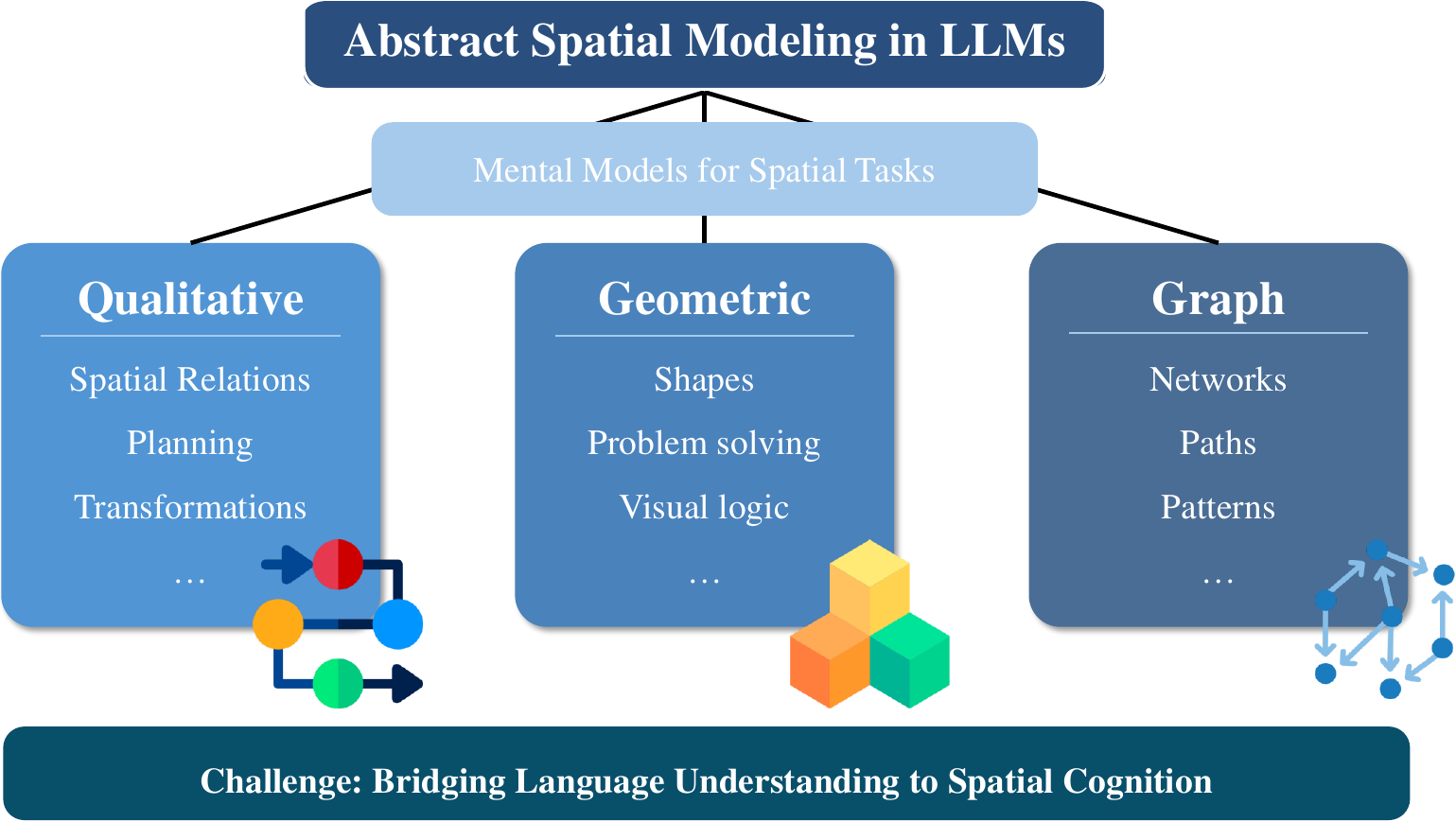}
    \caption{Conceptual framework of Abstract Spatial Reasoning. The framework illustrates three primary dimensions of spatial reasoning capabilities: qualitative reasoning, geometric reasoning, and graph reasoning. LLMs still face the challenge of bridging language understand to abstract spatial cognition.}
    \label{fig:abs_reason}
\end{figure*}

\section{LLM based Spatial Intelligence for the Real World}

\subsection{Embodied Spatial Intelligence} 
\begin{figure*}
    \centering
    \includegraphics[width=0.7\linewidth]{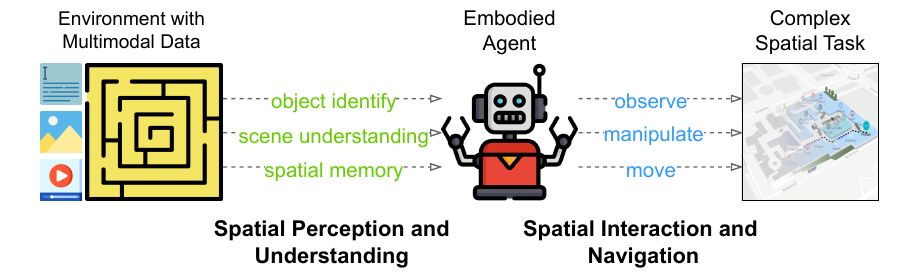}
    \caption{A simple schematic of embodied spatial intelligence. The framework illustrates two sequential stages: spatial perception and understanding and spatial interaction and navigation.}
    \label{fig:emb_spa_intell}
\end{figure*}

As shown in Fig.~\ref{fig:emb_spa_intell}, spatial intelligence in embodied AI comprises two key stages: 1) spatial perception and understanding, where agents acquire and process spatial information to construct internal representations of the environment, and 2) spatial interaction and navigation, where these representations are leveraged for movement, task execution, and decision-making.

\subsubsection{Spatial Perception and Understanding}
Spatial perception and understanding are essential for embodied intelligence, allowing agents (e.g., robots) to navigate, interact, and reason about their surroundings. Recent research has explored how multi-modal large language models (MLLMs) enhance these capabilities by integrating visual and textual data, improving spatial reasoning, and enabling interactive decision-making. Advancements in this field primarily involve three aspects: multi-modal spatial perception, scene-level spatial reasoning, and memory-based spatial exploration.

Multimodal spatial perception focuses on fusing RGB, depth, and textual information to enhance object localization and understanding. For instance, LLMI3D~\cite{yang2024llmi3d} enables 3D object position estimation from a single 2D image using spatial-enhanced feature extraction and 3D query token-based decoding. SpatialBot~\cite{cai2024spatialbot} integrates depth perception to improve robotic manipulation and spatial reasoning, supported by its SpatialQA dataset, which trains models in depth estimation and object grounding. While these approaches expand LLMs’ perceptual abilities, challenges remain in effectively integrating multi-modal data and improving fine-grained depth reasoning.

Beyond object-level perception, scene-level spatial reasoning enables agents to understand spatial relationships, align multi-view information, and interpret dynamic environments. Video-3D LLM~\cite{zheng2024video} enhances video-based LLMs by embedding 3D spatial coordinates into video features, supporting 3D question answering, visual grounding, and dense captioning. Scene-LLM~\cite{fu2024scene} integrates egocentric and global 3D scene representations, using 3D point-based features for more effective scene understanding and interactive planning. These models improve agents’ ability to process spatial information over time, though aligning continuous 3D spatial structures with language-based reasoning remains an open challenge.

For long-term spatial reasoning and adaptive decision-making, memory-based spatial exploration allows agents to retain and recall spatial knowledge. For example, 3D-Mem~\cite{yang20243dmem} introduces multi-view Memory Snapshots to store explored spatial data and frontier snapshots to identify unexplored areas, helping agents balance knowledge retrieval and active exploration. This approach enhances lifelong learning and autonomous adaptation, yet ensuring scalability and developing efficient retrieval mechanisms will be important for practical deployment.

As research progresses, improving multi-modal fusion, refining spatial reasoning, and optimizing memory mechanisms will be crucial for advancing LLM-driven spatial perception and understanding in embodied intelligence.

\subsubsection{Spatial Interaction and Navigation}
Spatial interaction and navigation involve action execution based on spatial perception and understanding. The actions include planning robotic actions and predicting future trajectories in spatial environments. Emerging research has dived into combining MLLMs in spatial interaction and navigation. Progress in this area mainly focuses on two aspects: motion control and navigation. 

Motion control can be categorized into simple action generation and interaction with a complex environment. The former applies the perception ability of MLLMs to directly generate the target action. For example, RT-2 \cite{zitkovich2023rt} integrates vision-language models (VLMs) pre-trained on internet-scale data into robot actions generation. VIMA \cite{jiang2022vima} leverages a transformer-based architecture designed to process multimodal prompts and generate motor actions autoregressively. However, in a complex environment, the reasoning ability enables spatial intelligence to handle open-set tasks. VexPoser \cite{huang2023voxposer} generates 3D spatial representations and plan robot actions by leveraging MLLMs’ reasoning and code-writing capabilities. 
GAJ-VGG \cite{wang2023generating} designs a graph neural network (Graph Action Justification) to construct a graph data representing the layout of obstacles and their surrounding environment through spatial and semantic relationships, and the robot outputs the optimal action.

Navigation task perceives and memorizes the surrounding environment, and predict the next location through reasoning. Based on the category of large model employed, navigation can be divided into language-model-based and vision-language-model-based task.
By feeding structured text-based maps into an LLM, Guide-LLM \cite{song2024guide} achieves indoor spatial perception and leverages the reasoning capabilities of LLM for path planning. 
NavGPT \cite{zhou2024navgpt} perceives the environment by using vision models to convert environment images into text and applies an LLM to integrate the current environmental descriptions with historical environment summaries, and perform trajectory planning. 
To bridge the gap between LLM-based navigation paradigms and Vision-Language-Navigation(VLN)-specialized models, NavGPT-2 \cite{zhou2025navgpt} integrates indoor visual observation with MLLMs and combining navigation policy networks to improve navigational reasoning. 
TopV-Nav \cite{zhong2024topv} prompts MLLMs with the spatial arrangement of objects using bounding boxes and text labels in the bird-view environment image and conducts dynamic map scaling and target-guided navigation through MLLM reasoning. 
MP5 \cite{qin2024mp5} designs an embodied system that decomposes complex open-world tasks and perceives the environment through active perception in Minecraft by calling MLLMS. VSI-Bench \cite{yang2024thinking} probes the MLLMs to conduct indoor route planning and finds that MLLMs can work effectively with naive cognitive map design. 
NWM \cite{bar2024navigation} proposes a controllable video generation model that predicts future target frame for navigation.

\subsection{Urban Spatial Intelligence} 
\begin{figure*}
    \centering
    \includegraphics[width=0.75\linewidth]{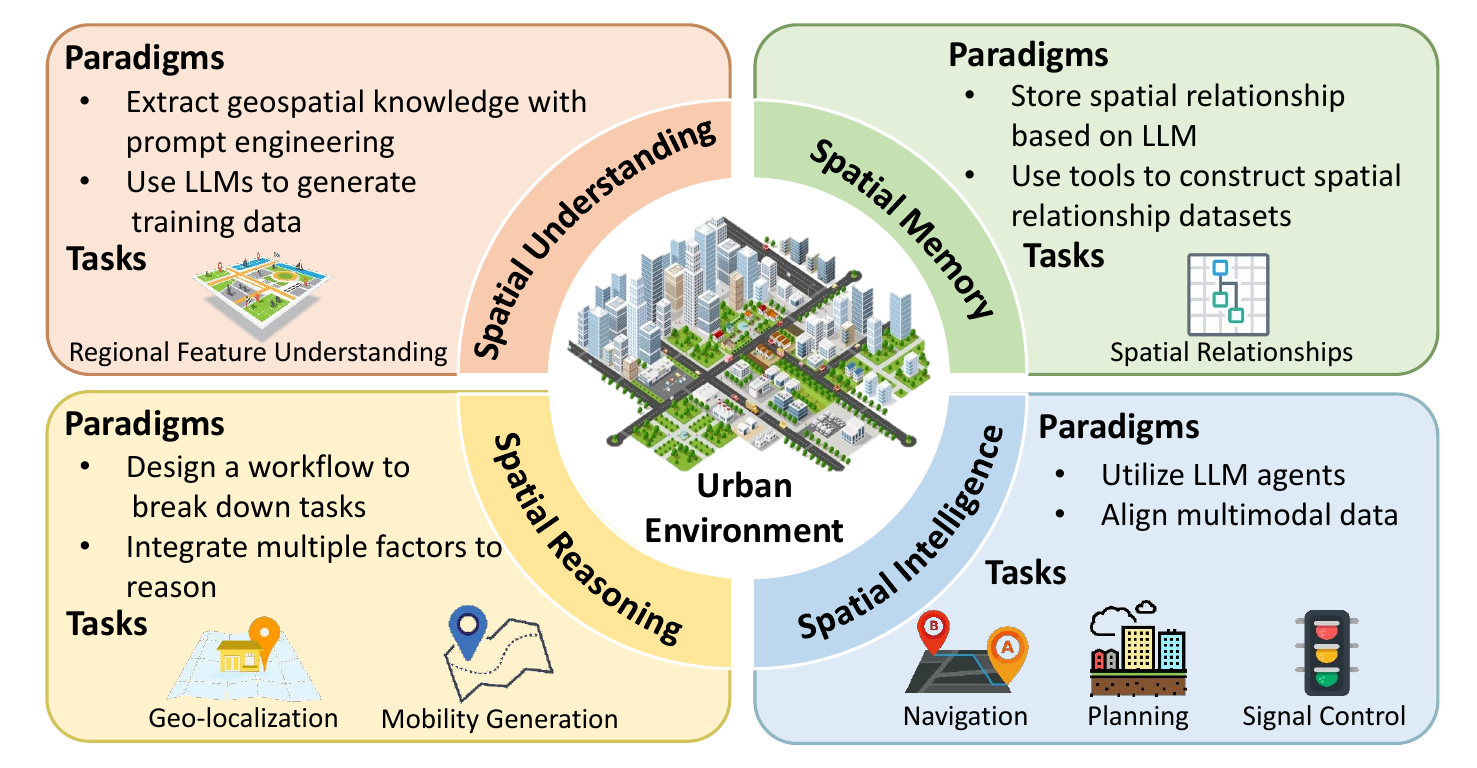}
    \caption{Urban spatial intelligence can be categorized into four main types: spatial understanding, spatial memory, spatial reasoning, and spatial intelligence. Each type includes its unique tasks and paradigms.}
    \label{fig:urban}
\end{figure*}
The embodied spatial intelligence primarily involves interaction and movement within arm's-reach micro-spaces, whereas at larger scales, LLMs necessitate fundamentally distinct spatial reasoning paradigms. This paradigm shift stems from a critical scaling effect: as spatial dimensions expand, the agent's physical size becomes negligible relative to the environment. Consequently, the agent transitions from operating within a body-embedded concrete space to processing extended spatial domains beyond immediate physical reach. This transformation necessitates a cognitive shift from subjective embodiment to objective spatial representation, requiring LLMs to conceptualize space as an independent entity with abstract properties. Such representational capacity enables advanced spatial functions including but not limited to cognitive mapping, pathfinding, trajectory optimization, and even generative spatial design.

Urban environments emerge as an optimal testing ground for these macro-scale spatial intelligence developments. As the most complex human-created spatial systems, cities integrate heterogeneous elements into multilayered structures encompassing physical infrastructure, functional zones, and socioeconomic networks. Their inherent spatial complexity has already propelled interdisciplinary research frontiers like urban computing and spatial econometrics, establishing essential methodological foundations. As shown in Figure~\ref{fig:urban}, to systematically investigate urban spatial intelligence, we propose a framework that distinguishes between understanding, memory, reasoning, and intelligence capabilities. The former evaluates the ability of LLMs to encode and retain massive urban elements, while the latter examines their operational competence in executing urban-specific tasks such as mobility simulation, service allocation optimization, and urban planning.

\subsubsection{Spatial Understanding and Memory} 
Spatial memory refers to the ability of models to recall geographic information and relationships between different spatial elements~\cite{gurnee2024language}.  Pre-trained large language models (LLMs) naturally acquire spatial priors from the geographical data embedded in their training corpus~\cite{manvi2024large}. This enables models to recognize, store, and retrieve spatial information in a way that mimics human spatial memory, which is crucial for tasks that require geographic reasoning or interpretation.

It can be categorized into two key aspects: (1) regional feature understanding and (2) reasoning about spatial locations and relationships.
To understand regional features, Manvi \textit{et al.}~\cite{manvi2023geollm} have proposed GEOLLM to extract geospatial knowledge from LLMs.  
The biases in geographic information learned by LLMs are also examined~\cite{manvi2024large}. 
Kuckreja \textit{et al.}\cite{kuckreja2024geochat} utilize satellite images to understand regional features. Satellite images, combined with LLMs, are also used to predict socioeconomic indicators~\cite{yan2024urbanclip}. Moreover, multimodal data—such as satellite images, language, and Points of Interest (POIs)—is employed to better understand regional characteristics and predict socioeconomic outcomes~\cite{xiao2024refound}. To reason about spatial locations and relationships, Ning \textit{et al.}\cite{ning2024urbankgent} leverage LLM-Agent to construct urban knowledge graphs~\cite{liu2022developing,liu2023urbankg}. 
We summarize the key methodologies for both aspects of spatial understanding. For regional feature understanding, one common approach is extracting prior knowledge through prompt engineering, which involves collecting spatial information from open-source data and aligning regional features using multimodal data integration. Another important strategy is leveraging LLMs to assist downstream tasks by generating training data and providing guidance for model training. Regarding spatial locations and relationships, models can infer spatial structures based on their pre-trained priors, using embedded geographic knowledge to reason about spatial relationships. Additionally, automated tools have been developed to construct and validate relationship datasets, facilitating the structured representation of spatial data and enhancing geographic reasoning.

\subsubsection{Spatial Reasoning and Intelligence} 
Spatial reasoning in cities refers to deriving new spatial information or predicting future urban dynamics based on spatial data or spatial relationships through reasoning. %
For example, GeoReasoner is a framework that integrates LLMs for geospatial localization, leveraging high-quality street view datasets to enhance spatial reasoning capabilities~\cite{ligeoreasoner}. %
Moreover, some research focuses on reasoning about the potential behavior patterns of urban residents. Wang et al. use LLM to model individual mobility in two stages: first, identifying spatiotemporal patterns of residents' mobility, and second, using these patterns to generate trajectories~\cite{wang2024large}. Similarly, Feng et al. break the trajectory prediction into three sub-tasks that influence mobility: remembering individual mobility patterns, learning shared spatial transition relationships of the group, and integrating spatial knowledge of urban structures, fully leveraging LLMs’ knowledge of geographic space~\cite{feng2024agentmove}. Shao et al. develop a Chain of Planned Behavior, which leverages the step-by-step reasoning capability of LLMs to achieve recursive inference of mobility intentions~\cite{shao2024beyond}. Gong et al. design a visiting intent memory network and a human travel preference prompt pool to help LLMs better understand the semantics of visiting intentions and travel preferences~\cite{gong2024mobility}.

Spatial intelligence in cities focuses on making decisions and responding based on spatial data, with the ability to make real-time judgments in complex urban environments. 
For example, urban planning is a typical task that requires spatial decision-making. Zhou et al. propose a multi-agent collaborative framework for participatory urban planning~\cite{zhou2024large}. %
Moreover, traffic signal control dynamically adjusts to the spatial environment, optimizing the traffic system's overall efficiency.
LLMLight integrates the task description and real-time traffic conditions into the prompt, leveraging the LLM's Chain-of-Thought reasoning capability to determine the optimal control strategy~\cite{lai2023large}.
Navigation tasks can recognize real-time changes in complex spatial environments, providing optimal navigation solutions.
For example, Xu et al. propose Flame~\cite{xu2024flame}, which enhances reasoning capabilities in three stages: from understanding a single street view description task to handling path planning tasks with multiple images, and ultimately achieving end-to-end spatial decision-making for navigation. Schumann et al. combine LLM with real-world environmental interaction, using a linguistic approach to process trajectories and visual observations, providing contextual prompts to the LLM to solve decision-making problems in navigation tasks~\cite{schumann2024velma}. Specifically, Zeng et al. propose a Perceive-Reflect-Plan workflow, enabling the LLM agent to autonomously navigate in urban environments~\cite{zeng2024perceive}.

\subsection{Earth Spatial Intelligence} 
\begin{figure*}
    \centering
    \includegraphics[width=0.95\linewidth]{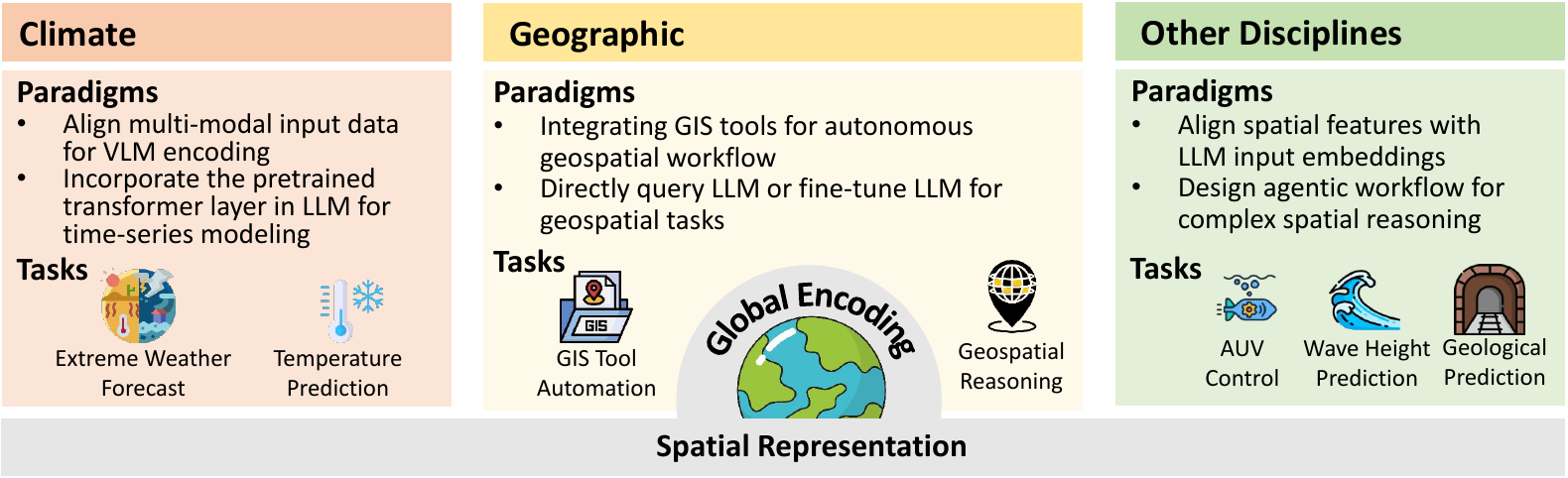}
    \caption{Illustrations of representative earth spatial intelligence fields and paradigms.}
    \label{fig:earth}
\end{figure*}

Earth Spatial Intelligence (ESI) is an interdisciplinary field at the intersection of artificial intelligence and Earth sciences. ESI addresses complex challenges across domains, including climate science, geography, oceanography, and geology, by leveraging large-scale spatio-temporal data and cutting-edge techniques like LLMs and multimodal LLMs (MLLMs). These models process vast datasets, uncover patterns, and generate insights that drive modeling, decision-making, and environmental resilience advancements. In climate science, LLMs enhance the forecasting of precipitation and climate events by capturing spatio-temporal dependencies and integrating meteorological raster data. In geography, they combine with Geographic Information Systems (GIS) for automated geospatial reasoning and localized spatial analyses while improving contextual deduction through adaptive modules and contrastive learning. In oceanography, vision-language models enable natural language control of Autonomous Underwater Vehicles (AUVs), while spatio-temporal encoding addresses data sparsity, advancing wave height prediction and marine environmental modeling. In geology, LLMs integrate imagery and surveys to model geological phenomena, improve spatial reasoning, and streamline remote sensing-based mineral exploration. ESI is transforming Earth sciences by uniting natural language understanding, multimodal integration, and spatio-temporal reasoning. This rapidly evolving field offers profound opportunities for scientific discovery, sustainable resource management, and tackling pressing global challenges.

\subsubsection{Global Encoding}
At the global scale, a crucial aspect of intelligence is the proper encoding of location, enabling machines to perceive and understand spatial information effectively. While large language model-based applications typically represent location using longitude and latitude~\cite{manvi2023geollm,yan2024georeasoner}, machine learning and deep learning approaches have adopted a variety of spatial representation methods~\cite{wu2024torchspatial}. Specifically, 2D representation methods include approaches such as direct tile ID encoding, sinusoidal location encoders, and kernel-based techniques, while 3D methods encompass Cartesian coordinate encoding and various self-supervised representation strategies. According to the TorchSpatial benchmark~\cite{wu2024torchspatial}, the Sphere2Vec-sphereC+ method~\cite{mai2023sphere2vec}---a self-supervised 3D encoding technique that preserves the order between any two points on Earth---is the most effective and informative location encoding approach. Notably, even the direct tile ID encoding method—despite being the lowest-performing among common spatial representation techniques—significantly outperforms GPT-4V~\cite{wu2024torchspatial}. This phenomenon may underscore the discouraging applicability of large language models to explicit spatial learning tasks; however, they excel in few-shot, zero-shot, and similar scenarios, and demonstrate remarkable flexibility in leveraging multi-source data.

\subsubsection{Climate}
Climate events have a strong spatio-temporal dependency, which has been summarized as knowledge and commanded by language models to some extent. Therefore, there has been some trials in utilizing language models to predict or forecast climate events. LLMDiff incorporated a frozen transformer block from pre-trained LLM to serve as a universal visual encoder layer, with an intention of capturing long-term temporal dependencies and accurately estimating motion trends for improved precipitation nowcasting~\cite{she2024llmdiff}. CLLMate incorporated LLM and VLM to align meteorological raster data with weather and climate event information and train on the aligned datasets, enabling accurate forecasting of climate events with raster data~\cite{li2024cllmate}.
Notably, for the climate domain, large models have been largely applied and explored. GenCast~\cite{ravuri2021skilful} proposed a machine learning-based weather prediction model that generates accurate 15-day probabilistic ensemble weather forecasts. Pangu-Weather~\cite{bi2023accurate} introduced three-dimensional deep networks with Earth-specific priors and a hierarchical temporal aggregation strategy to achieve medium-range global weather forecasting. NowcastNet~\cite{zhang2023skilful} achieved nonlinear nowcasting for extreme precipitation by combining physical-evolution schemes and conditional-learning methods to produce high-resolution, physically plausible forecasts with lead times up to 3 hours. Fuxi~\cite{chen2023fuxi} introduced a cascaded machine learning weather forecasting system, which utilizes 39 years of ECMWF ERA5 reanalysis data to provide 15-day global forecasts at a 6-hour temporal resolution and 0.25° spatial resolution. The success of large models in climate modeling validates the growing prediction capabilities through training with large-scale data.

\subsubsection{Geography }
Considering the rich geographic knowledge commanded by large language models, their direct application to geography-related tasks has been widely explored. Geography-related tasks either involve the extraction and sensing of location-related knowledge across the global scale, or tasks requiring direct judgments and operations involving specific locations, such as localization and mapping. Two benchmark works comprehensively assess large language models' capacities in these two types of tasks. Manvi et al. find that naively querying LLMs using geographic coordinates alone is ineffective for predicting key indicators like population density; however, incorporating auxiliary map data from OpenStreetMap into the prompts significantly improves prediction accuracy~\cite{manvi2023geollm}. Roberts et al. find that while MLLMs perform well in memory-based geographic tasks, such as identifying locations or recognizing patterns from provided information, they face significant challenges in reasoning-based or more intelligent tasks, such as contextual deduction and advanced geospatial analysis~\cite{roberts2024charting}. To address the existing limitations of large language models, GeoGPT utilizes mature GIS tools to tackle geospatial tasks, integrating the semantic understanding ability of LLMs with GIS tools in an autonomous manner~\cite{zhang2023geogpt}. GeoSEE incorporates six information collection modules, which LLMs automatically select to adapt to specific indicators and countries~\cite{han2024geosee}. GeoReasoner incorporates two contrastive losses to enhance the reasoning ability of language models by making representations of nearby locations and the same entities more similar~\cite{yan2024georeasoner}.

\subsubsection{Other Disciplines}
LLMs have also been applied in other disciplines such as marine science and geology. With remarkable abilities like natural language understanding, generalizability, and reasoning, LLMs have been leveraged to tackle typical challenges in these disciplines such as data sparsity and complex decision-making.

In marine science, LLMs have been used for vehicle control due to their capability of spatial planning and reasoning. For example, OceanPlan leverages LLMs to control Autonomous Underwater Vehicle (AUV) through natural language command~\cite{yang2024oceanplan}. Specifically, it leverages a vision-language model to convert image observation into textual semantic map to memorize the explored ocean environment. It further proposes a hierarchical planning framework to convert natural language commands to control inputs for AUV, and adaptively adjust the plan in special circumstances.
Moreover, the generalization and few-shot learning abilities of LLMs are suitable for addressing the data-sparsity issue in spatial prediction. Li et al. use LLMs to predict the ocean significant wave height with sparse observation data~\cite{li2024ocean}. To enhance the spatial understanding ability of LLM, they first encode the spatio-temporal features from the observation data through a spatio-temporal encoder, which is then aligned with the embeddings of natural language prompt and fed into the LLM together for prediction.

In geology, Xu et al. use LLMs to predict the geological condition in tunnels~\cite{xu2024geopredict}. They first construct a knowledge graph (KG) to integrate multimodal data and transform them into low-dimensional KG embeddings. Then they align the KG embeddings with prompt embeddings through patch reprogramming, and input them into LLM for prediction.
Yu et al. propose a multi-agent collaboration framework to enhance the spatial reasoning ability of MLLM for remote-sensing mineral exploration~\cite{yu2024mineagent}. It construct multiple MLLM agents responsible for identifying different features from different remote-sensing images and integrate them together, which shows considerable performance. 

Overall, the application of LLM spatial intelligence in these disciplines can be summarized in two ways: (1) Aligning spatial features with prompt embeddings and input them into the LLM for prediction tasks. (2) Designing agentic workflow with LLMs to enable complex spatial reasoning.

\section{Challenges and Discussions}

\subsection{Fundamental Spatial Intelligence} 
The study of fundamental spatial intelligence raises several critical questions and challenges. First, the form of spatial reasoning—the core of spatial intelligence—remains a central issue: is language-based spatial reasoning the most effective form currently known, or are there more universal and effective modeling approaches, such as graph-based representations or multi-modal frameworks? Second, the comprehensive evaluation of general spatial intelligence poses a significant challenge. Current frameworks often focus on specific tasks or domains, lacking a unified approach to assess spatial intelligence across diverse contexts, domains, and scales. Such a unified evaluation is crucial for understanding the relationship between fundamental spatial intelligence and its manifestations in other domains. This requires investigating how core spatial abilities, like mental rotation or spatial memory, translate into higher-order applications in specialized fields. Addressing these questions will not only advance our theoretical understanding of spatial intelligence but also inform the development of more robust and effective models for artificial general intelligence.

\subsection{Embodied Spatial Intelligence} 
For embodied intelligence, two significant challenges remain in the research on spatial memory and intelligence.
First, the current work on embodied intelligence only partially incorporates prior knowledge of spatial cognition as a source of inspiration in method design. While some studies draw loosely from principles of human spatial cognition—such as wayfinding, mental mapping, or object manipulation—these inspirations are often superficial and lack a systematic integration into the computational models.  Therefore, there is a pressing need for an approach that deeply couples model design with the underlying mechanisms of human spatial cognition. Such an approach would not only improve the robustness and adaptability of these models but also provide insights into the fundamental principles of human intelligence. However, achieving this integration is inherently challenging, as it requires bridging the gap between cognitive science, neuroscience, and embodied artificial intelligence.
Second, research on embodied intelligence encompasses a wide spectrum of multi-level spatial intelligence and cognition, each with distinct characteristics. For example, at the lower level, tasks such as robotic manipulation demand fine-grained motor control and precise spatial reasoning to interact with objects in a constrained environment. On the other hand, higher-level tasks like path-planning for unmanned aerial vehicles (UAVs) involve large-scale spatial reasoning.  Therefore, it is an open question whether it is possible to build a universal model integrating multi-level (\textit{i.e.},  multi-grained) spatial intelligence in embodied AI tasks.

\subsection{Urban Spatial Intelligence} 
Although significant progress has been made in urban spatial intelligence, several critical challenges remain. First, the heterogeneity of urban data poses fundamental limitations: current frameworks struggle to harmonize multimodal inputs (e.g., satellite imagery, POIs, and mobility patterns) into unified spatial representations, often leading to fragmented understanding. And the most often text-based representation of complex spatial structures is always doubtable for urban professionals. Second, the robustness of spatial reasoning remains constrained by LLMs' reliance on static training data, which inadequately capture dynamic urban phenomena such as real-time traffic flows or evolving socioeconomic factors. Third, the interpretability gap in LLM-driven spatial decisions in urban planning and navigation tasks raises concerns about trustworthiness, particularly when models prioritize statistical correlations over causal spatial relationships. Therefore, future research may prioritize three directions: (1) dynamic spatial modeling to integrate real-time data with LLMs, enabling adaptive responses to urban dynamics while addressing constraints; (2) Causal spatial reasoning frameworks that disentangle environmental, social, and infrastructural interdependencies, solving the concern and resistance about dealing spatial information in text paradigm; (3) Ethical challenges in the mitigation of spatial bias, which is highlighted by geographic priors in LLM, demand systematic auditing methods to ensure equitable urban intelligence applications.

\subsection{Earth Spatial Intelligence} 
LLM holds transformative potential for advancing Earth Spatial Intelligence, but several challenges must be overcome to fully realize their capabilities. One key limitation is their performance in reasoning-intensive tasks, such as contextual deduction and advanced spatial analysis in geography, geology, and other domains, where bottlenecks persist. While multimodal LLMs (MLLMs) and emerging frameworks like GeoReasoner and MineAgent show promise by leveraging contrastive learning and multi-agent systems, further innovation is required to achieve robust geospatial understanding. The integration of domain-specific data also presents significant hurdles. For instance, marine sciences often grapple with data sparsity, necessitating tailored solutions like OceanGPT and spatio-temporal encoders. Meanwhile, domains like geology and climate science depend heavily on complex and multimodal inputs, including knowledge graph embeddings and specialized prompts, which demand seamless alignment within LLM architectures.  Future research directions include leveraging transfer learning to adapt pre-trained models across related Earth science domains, thereby reducing data requirements and fostering knowledge sharing. Benchmarking platforms like OceanBench and integrated systems such as GeoGPT could provide standardization and rigorous evaluation across ESI subfields, enabling targeted advancements. Human-in-the-loop systems and explainable AI (XAI) frameworks could further enhance interpretability and trust, while advances in causal inference offer the potential to better capture dynamic Earth processes. Interdisciplinary collaboration will be essential to translate these advancements into actionable solutions for climate resilience and sustainable development. By tackling these challenges, LLMs can unlock more precise predictions and insights to address global environmental challenges.

\subsection{Relation with World Model}
In this paper, we investigate spatial understanding and task-solving within the domain of spatial intelligence. The concept of world models has recently emerged as a significant topic in this field, particularly in embodied spatial intelligence, propelled by advancements in diffusion-based generative models. As outlined in a recent survey~\cite{ding2024understanding}, world models—rooted in psychological mental models—serve two key functions: constructing internal representations to interpret the underlying mechanisms of the world and predicting future states to guide decision-making. Our work primarily focuses on the first function, developing internal representations to deepen spatial comprehension. In computational terms, this aligns with model-based reinforcement learning, where parameterized environmental models enhance intelligent behavior. While we address most aspects of world models, our emphasis lies in understanding rather than the generative aspect, such as forecasting outcomes. For a more extensive exploration of generative capabilities, we refer readers to~\cite{ding2024understanding}. Looking forward, we propose that integrating these generative capabilities into spatial intelligence modeling holds considerable promise. This could enable more robust systems capable of not only understanding but also predicting and acting within the physical world, potentially addressing limitations seen in current foundation models, such as the lack of granularity in urban knowledge highlighted by Feng et al.~\cite{feng2024citygpt,feng2024citybench}.

\section{Conclusion}
This paper begins with a discussion of human spatial intelligence research in neuroscience and cognitive science, reviewing and summarizing studies on spatial intelligence across various disciplines, particularly at different spatial scales, since the era of LLMs. It aims to provide a comprehensive overview of spatial intelligence research across domains, helping to contextualize existing studies and inspire future research directions. We believe that cross-domain spatial intelligence research at multi-scales will emerge as a crucial area of study in the future, generating significant impacts and profound applications across multiple fields. Furthermore, in-depth investigations into spatial intelligence will, in turn, inform the development of general artificial intelligence, laying a solid foundation for humanity's advancement toward true artificial general intelligence.

\newpage

\bibliographystyle{named}
\bibliography{ijcai25_origin}

\begin{thebibliography}{}

\bibitem[\protect\citeauthoryear{Bar \bgroup \em et al.\egroup }{2024}]{bar2024navigation}
Amir Bar, Gaoyue Zhou, Danny Tran, Trevor Darrell, and Yann LeCun.
\newblock Navigation world models.
\newblock {\em arXiv preprint arXiv:2412.03572}, 2024.

\bibitem[\protect\citeauthoryear{Bhandari \bgroup \em et al.\egroup }{2023}]{bhandari2023large}
Prabin Bhandari, Antonios Anastasopoulos, and Dieter Pfoser.
\newblock Are large language models geospatially knowledgeable?
\newblock In {\em Proceedings of the 31st ACM International Conference on Advances in Geographic Information Systems}, pages 1--4, 2023.

\bibitem[\protect\citeauthoryear{Bi \bgroup \em et al.\egroup }{2023}]{bi2023accurate}
Kaifeng Bi, Lingxi Xie, Hengheng Zhang, Xin Chen, Xiaotao Gu, and Qi~Tian.
\newblock Accurate medium-range global weather forecasting with 3d neural networks.
\newblock {\em Nature}, 619(7970):533--538, 2023.

\bibitem[\protect\citeauthoryear{Cai \bgroup \em et al.\egroup }{2024}]{cai2024spatialbot}
Wenxiao Cai, Yaroslav Ponomarenko, Jianhao Yuan, Xiaoqi Li, Wankou Yang, Hao Dong, and Bo~Zhao.
\newblock Spatialbot: Precise spatial understanding with vision language models.
\newblock {\em arXiv preprint arXiv:2406.13642}, 2024.

\bibitem[\protect\citeauthoryear{Chen \bgroup \em et al.\egroup }{2023}]{chen2023fuxi}
Lei Chen, Xiaohui Zhong, et~al.
\newblock {F}uxi: {A} cascade machine learning forecasting system for 15-day global weather forecast.
\newblock {\em npj Clim. Atmos. Sci.}, 2023.

\bibitem[\protect\citeauthoryear{Chen \bgroup \em et al.\egroup }{2024}]{chen2024spatialvlm}
Boyuan Chen, Zhuo Xu, et~al.
\newblock {S}patialvlm: {E}ndowing vision-language models with spatial reasoning capabilities.
\newblock In {\em Proc. of CVPR}, 2024.

\bibitem[\protect\citeauthoryear{Cohen}{1993}]{cohen1993memory}
NJ~Cohen.
\newblock {\em Memory, amnesia and the hippocampal system}.
\newblock MIT Press, 1993.

\bibitem[\protect\citeauthoryear{Ding \bgroup \em et al.\egroup }{2024}]{ding2024understanding}
Jingtao Ding, Yunke Zhang, Yu~Shang, Yuheng Zhang, Zefang Zong, Jie Feng, Yuan Yuan, Hongyuan Su, Nian Li, Nicholas Sukiennik, et~al.
\newblock Understanding world or predicting future? a comprehensive survey of world models.
\newblock {\em arXiv preprint arXiv:2411.14499}, 2024.

\bibitem[\protect\citeauthoryear{Eichenbaum and Cohen}{2014}]{eichenbaum2014can}
Howard Eichenbaum and Neal~J Cohen.
\newblock Can we reconcile the declarative memory and spatial navigation views on hippocampal function?
\newblock {\em Neuron}, 83(4):764--770, 2014.

\bibitem[\protect\citeauthoryear{Epstein \bgroup \em et al.\egroup }{2017}]{epstein2017cognitive}
Russell~A Epstein, Eva~Zita Patai, Joshua~B Julian, and Hugo~J Spiers.
\newblock The cognitive map in humans: spatial navigation and beyond.
\newblock {\em Nature neuroscience}, 20(11):1504--1513, 2017.

\bibitem[\protect\citeauthoryear{Farzanfar \bgroup \em et al.\egroup }{2023}]{farzanfar2023cognitive}
Delaram Farzanfar, Hugo~J Spiers, Morris Moscovitch, and R~Shayna Rosenbaum.
\newblock From cognitive maps to spatial schemas.
\newblock {\em Nature Reviews Neuroscience}, 24(2):63--79, 2023.

\bibitem[\protect\citeauthoryear{Feng \bgroup \em et al.\egroup }{2024a}]{feng2024citygpt}
Jie Feng, Yuwei Du, Tianhui Liu, Siqi Guo, Yuming Lin, and Yong Li.
\newblock Citygpt: Empowering urban spatial cognition of large language models.
\newblock {\em arXiv preprint arXiv:2406.13948}, 2024.

\bibitem[\protect\citeauthoryear{Feng \bgroup \em et al.\egroup }{2024b}]{feng2024agentmove}
Jie Feng, Yuwei Du, Jie Zhao, and Yong Li.
\newblock Agentmove: Predicting human mobility anywhere using large language model based agentic framework.
\newblock {\em arXiv preprint arXiv:2408.13986}, 2024.

\bibitem[\protect\citeauthoryear{Feng \bgroup \em et al.\egroup }{2024c}]{feng2024citybench}
Jie Feng, Jun Zhang, Tianhui Liu, Xin Zhang, Tianjian Ouyang, Junbo Yan, Yuwei Du, Siqi Guo, and Yong Li.
\newblock Citybench: Evaluating the capabilities of large language models for urban tasks, 2024.

\bibitem[\protect\citeauthoryear{Fu \bgroup \em et al.\egroup }{2024}]{fu2024scene}
Rao Fu, Jingyu Liu, Xilun Chen, Yixin Nie, and Wenhan Xiong.
\newblock Scene-llm: Extending language model for 3d visual understanding and reasoning.
\newblock {\em arXiv preprint arXiv:2403.11401}, 2024.

\bibitem[\protect\citeauthoryear{Gilboa and Marlatte}{2017}]{gilboa2017neurobiology}
Asaf Gilboa and Hannah Marlatte.
\newblock Neurobiology of schemas and schema-mediated memory.
\newblock {\em Trends in cognitive sciences}, 21(8):618--631, 2017.

\bibitem[\protect\citeauthoryear{Gong \bgroup \em et al.\egroup }{2024}]{gong2024mobility}
Letian Gong, Yan Lin, Xinyue Zhang, Yiwen Lu, Xuedi Han, Yichen Liu, Shengnan Guo, Youfang Lin, and Huaiyu Wan.
\newblock Mobility-llm: Learning visiting intentions and travel preferences from human mobility data with large language models.
\newblock {\em arXiv preprint arXiv:2411.00823}, 2024.

\bibitem[\protect\citeauthoryear{Gupta \bgroup \em et al.\egroup }{2021}]{gupta2021embodied}
Agrim Gupta, Silvio Savarese, et~al.
\newblock {E}mbodied intelligence via learning and evolution.
\newblock {\em Nature communications}, 2021.

\bibitem[\protect\citeauthoryear{Gurnee and Tegmark}{2024}]{gurnee2024language}
Wes Gurnee and Max Tegmark.
\newblock Language models represent space and time, 2024.

\bibitem[\protect\citeauthoryear{Haas \bgroup \em et al.\egroup }{2024}]{haas2024pigeon}
Lukas Haas, Michal Skreta, Silas Alberti, and Chelsea Finn.
\newblock Pigeon: Predicting image geolocations.
\newblock In {\em Proceedings of the IEEE/CVF Conference on Computer Vision and Pattern Recognition}, pages 12893--12902, 2024.

\bibitem[\protect\citeauthoryear{Han \bgroup \em et al.\egroup }{2024}]{han2024geosee}
Sungwon Han, Donghyun Ahn, Seungeon Lee, Minhyuk Song, Sungwon Park, Sangyoon Park, Jihee Kim, and Meeyoung Cha.
\newblock Geosee: Regional socio-economic estimation with a large language model.
\newblock {\em arXiv preprint arXiv:2406.09799}, 2024.

\bibitem[\protect\citeauthoryear{Huang \bgroup \em et al.\egroup }{2023a}]{huang2023survey}
Lei Huang, Weijiang Yu, Weitao Ma, Weihong Zhong, Zhangyin Feng, Haotian Wang, Qianglong Chen, Weihua Peng, Xiaocheng Feng, Bing Qin, et~al.
\newblock A survey on hallucination in large language models: Principles, taxonomy, challenges, and open questions.
\newblock {\em arXiv preprint arXiv:2311.05232}, 2023.

\bibitem[\protect\citeauthoryear{Huang \bgroup \em et al.\egroup }{2023b}]{huang2023voxposer}
Wenlong Huang, Chen Wang, Ruohan Zhang, Yunzhu Li, Jiajun Wu, and Li~Fei-Fei.
\newblock Voxposer: Composable 3d value maps for robotic manipulation with language models.
\newblock {\em arXiv preprint arXiv:2307.05973}, 2023.

\bibitem[\protect\citeauthoryear{Ishikawa}{2021}]{ishikawa2021spatial}
Toru Ishikawa.
\newblock Spatial thinking, cognitive mapping, and spatial awareness.
\newblock {\em Cognitive Processing}, 22(Suppl 1):89--96, 2021.

\bibitem[\protect\citeauthoryear{Jiang \bgroup \em et al.\egroup }{2022}]{jiang2022vima}
Yunfan Jiang, Agrim Gupta, Zichen Zhang, Guanzhi Wang, Yongqiang Dou, Yanjun Chen, Li~Fei-Fei, Anima Anandkumar, Yuke Zhu, and Linxi Fan.
\newblock Vima: General robot manipulation with multimodal prompts.
\newblock {\em arXiv preprint arXiv:2210.03094}, 2(3):6, 2022.

\bibitem[\protect\citeauthoryear{Kazemi \bgroup \em et al.\egroup }{2023}]{kazemi2023geomverse}
Mehran Kazemi, Hamidreza Alvari, Ankit Anand, Jialin Wu, Xi~Chen, and Radu Soricut.
\newblock Geomverse: A systematic evaluation of large models for geometric reasoning.
\newblock {\em arXiv preprint arXiv:2312.12241}, 2023.

\bibitem[\protect\citeauthoryear{Kuckreja \bgroup \em et al.\egroup }{2024}]{kuckreja2024geochat}
Kartik Kuckreja, Muhammad~Sohail Danish, Muzammal Naseer, Abhijit Das, Salman Khan, and Fahad~Shahbaz Khan.
\newblock Geochat: Grounded large vision-language model for remote sensing.
\newblock In {\em Proceedings of the IEEE/CVF Conference on Computer Vision and Pattern Recognition}, pages 27831--27840, 2024.

\bibitem[\protect\citeauthoryear{Lai \bgroup \em et al.\egroup }{2023}]{lai2023large}
Siqi Lai, Zhao Xu, Weijia Zhang, Hao Liu, and Hui Xiong.
\newblock Large language models as traffic signal control agents: Capacity and opportunity.
\newblock {\em arXiv preprint arXiv:2312.16044}, 2023.

\bibitem[\protect\citeauthoryear{Lee \bgroup \em et al.\egroup }{2022}]{lee2022factuality}
Nayeon Lee, Wei Ping, Peng Xu, Mostofa Patwary, Pascale~N Fung, Mohammad Shoeybi, and Bryan Catanzaro.
\newblock Factuality enhanced language models for open-ended text generation.
\newblock {\em Advances in Neural Information Processing Systems}, 35:34586--34599, 2022.

\bibitem[\protect\citeauthoryear{Lehnert \bgroup \em et al.\egroup }{2024}]{lehnert2024beyond}
Lucas Lehnert, Sainbayar Sukhbaatar, DiJia Su, Qinqing Zheng, Paul Mcvay, Michael Rabbat, and Yuandong Tian.
\newblock Beyond a*: Better planning with transformers via search dynamics bootstrapping.
\newblock {\em arXiv preprint arXiv:2402.14083}, 2024.

\bibitem[\protect\citeauthoryear{Li \bgroup \em et al.\egroup }{2024a}]{li2024advancing}
Fangjun Li, David~C Hogg, and Anthony~G Cohn.
\newblock Advancing spatial reasoning in large language models: An in-depth evaluation and enhancement using the stepgame benchmark.
\newblock In {\em Proceedings of the AAAI Conference on Artificial Intelligence}, volume~38, pages 18500--18507, 2024.

\bibitem[\protect\citeauthoryear{Li \bgroup \em et al.\egroup }{2024b}]{li2024cllmate}
Haobo Li, Zhaowei Wang, Jiachen Wang, Alexis Kai~Hon Lau, and Huamin Qu.
\newblock Cllmate: A multimodal llm for weather and climate events forecasting.
\newblock {\em arXiv preprint arXiv:2409.19058}, 2024.

\bibitem[\protect\citeauthoryear{Li \bgroup \em et al.\egroup }{2024c}]{ligeoreasoner}
Ling Li, Yu~Ye, Bingchuan Jiang, and Wei Zeng.
\newblock Georeasoner: Geo-localization with reasoning in street views using a large vision-language model.
\newblock In {\em Forty-first International Conference on Machine Learning}, 2024.

\bibitem[\protect\citeauthoryear{Li \bgroup \em et al.\egroup }{2024d}]{li2024ocean}
Zhe Li, Ronghui Xu, Jilin Hu, Zhong Peng, Xi~Lu, Chenjuan Guo, and Bin Yang.
\newblock Ocean significant wave height estimation with spatio-temporally aware large language models.
\newblock In {\em Proceedings of the 33rd ACM International Conference on Information and Knowledge Management}, pages 3892--3896, 2024.

\bibitem[\protect\citeauthoryear{Li \bgroup \em et al.\egroup }{2024e}]{li2024urbangpt}
Zhonghang Li, Lianghao Xia, et~al.
\newblock {U}rbangpt: {S}patio-temporal large language models.
\newblock In {\em Proc. of KDD}, 2024.

\bibitem[\protect\citeauthoryear{Lin \bgroup \em et al.\egroup }{2024}]{lin2024advancese}
Jinzhou Lin, Han Gao, et~al.
\newblock {A}dvances in {E}mbodied {N}avigation {U}sing {L}arge {L}anguage {M}odels: {A} survey.
\newblock {\em arXiv:2311.00530}, 2024.

\bibitem[\protect\citeauthoryear{Liu \bgroup \em et al.\egroup }{2022}]{liu2022developing}
Yu~Liu, Jingtao Ding, and Yong Li.
\newblock Developing knowledge graph based system for urban computing.
\newblock In {\em Proceedings of the 1st ACM SIGSPATIAL International Workshop on Geospatial Knowledge Graphs}, pages 3--7, 2022.

\bibitem[\protect\citeauthoryear{Liu \bgroup \em et al.\egroup }{2023}]{liu2023urbankg}
Yu~Liu, Jingtao Ding, Yanjie Fu, and Yong Li.
\newblock Urbankg: An urban knowledge graph system.
\newblock {\em ACM Transactions on Intelligent Systems and Technology}, 14(4):1--25, 2023.

\bibitem[\protect\citeauthoryear{Long \bgroup \em et al.\egroup }{2025}]{long2025allocentric}
Xiaoyang Long, Daniel Bush, Bin Deng, Neil Burgess, and Sheng-Jia Zhang.
\newblock Allocentric and egocentric spatial representations coexist in rodent medial entorhinal cortex.
\newblock {\em Nature Communications}, 16(1):356, 2025.

\bibitem[\protect\citeauthoryear{Luo \bgroup \em et al.\egroup }{2024}]{luo2024graphinstruct}
Zihan Luo, Xiran Song, Hong Huang, Jianxun Lian, Chenhao Zhang, Jinqi Jiang, and Xing Xie.
\newblock Graphinstruct: Empowering large language models with graph understanding and reasoning capability.
\newblock {\em arXiv preprint arXiv:2403.04483}, 2024.

\bibitem[\protect\citeauthoryear{Mai \bgroup \em et al.\egroup }{2021}]{mai2021geographic}
Gengchen Mai, Krzysztof Janowicz, Rui Zhu, Ling Cai, and Ni~Lao.
\newblock Geographic question answering: challenges, uniqueness, classification, and future directions.
\newblock {\em AGILE: GIScience series}, 2:8, 2021.

\bibitem[\protect\citeauthoryear{Mai \bgroup \em et al.\egroup }{2023}]{mai2023sphere2vec}
Gengchen Mai, Yao Xuan, et~al.
\newblock {S}phere2{V}ec: {A} general-purpose location representation learning over a spherical surface for large-scale geospatial predictions.
\newblock {\em ISPRS J. P. Remote Sens.}, 2023.

\bibitem[\protect\citeauthoryear{Mansourian and Oucheikh}{2024}]{mansourian2024chatgeoai}
Ali Mansourian and Rachid Oucheikh.
\newblock Chatgeoai: Enabling geospatial analysis for public through natural language, with large language models.
\newblock {\em ISPRS International Journal of Geo-Information}, 13(10):348, 2024.

\bibitem[\protect\citeauthoryear{Manvi \bgroup \em et al.\egroup }{2023}]{manvi2023geollm}
Rohin Manvi, Samar Khanna, Gengchen Mai, Marshall Burke, David Lobell, and Stefano Ermon.
\newblock Geollm: Extracting geospatial knowledge from large language models.
\newblock {\em arXiv preprint arXiv:2310.06213}, 2023.

\bibitem[\protect\citeauthoryear{Manvi \bgroup \em et al.\egroup }{2024}]{manvi2024large}
Rohin Manvi, Samar Khanna, Marshall Burke, David Lobell, and Stefano Ermon.
\newblock Large language models are geographically biased.
\newblock {\em arXiv preprint arXiv:2402.02680}, 2024.

\bibitem[\protect\citeauthoryear{Momennejad \bgroup \em et al.\egroup }{2024}]{momennejad2024evaluating}
Ida Momennejad, Hosein Hasanbeig, Felipe Vieira~Frujeri, Hiteshi Sharma, Nebojsa Jojic, Hamid Palangi, Robert Ness, and Jonathan Larson.
\newblock Evaluating cognitive maps and planning in large language models with cogeval.
\newblock {\em Advances in Neural Information Processing Systems}, 36, 2024.

\bibitem[\protect\citeauthoryear{Moser \bgroup \em et al.\egroup }{2008}]{moser2008place}
Edvard~I Moser, Emilio Kropff, and May-Britt Moser.
\newblock Place cells, grid cells, and the brain's spatial representation system.
\newblock {\em Annu. Rev. Neurosci.}, 31(1):69--89, 2008.

\bibitem[\protect\citeauthoryear{Moser \bgroup \em et al.\egroup }{2017}]{moser2017spatial}
Edvard~I Moser, May-Britt Moser, and Bruce~L McNaughton.
\newblock Spatial representation in the hippocampal formation: a history.
\newblock {\em Nature neuroscience}, 20(11):1448--1464, 2017.

\bibitem[\protect\citeauthoryear{Ning and Liu}{2024}]{ning2024urbankgent}
Yansong Ning and Hao Liu.
\newblock Urbankgent: A unified large language model agent framework for urban knowledge graph construction.
\newblock {\em arXiv preprint arXiv:2402.06861}, 2024.

\bibitem[\protect\citeauthoryear{Petroni \bgroup \em et al.\egroup }{2019}]{petroni2019language}
Fabio Petroni, Tim Rockt{\"a}schel, Patrick Lewis, Anton Bakhtin, Yuxiang Wu, Alexander~H Miller, and Sebastian Riedel.
\newblock Language models as knowledge bases?
\newblock {\em arXiv preprint arXiv:1909.01066}, 2019.

\bibitem[\protect\citeauthoryear{Qin \bgroup \em et al.\egroup }{2024}]{qin2024mp5}
Yiran Qin, Enshen Zhou, Qichang Liu, Zhenfei Yin, Lu~Sheng, Ruimao Zhang, Yu~Qiao, and Jing Shao.
\newblock Mp5: A multi-modal open-ended embodied system in minecraft via active perception.
\newblock In {\em 2024 IEEE/CVF Conference on Computer Vision and Pattern Recognition (CVPR)}, pages 16307--16316. IEEE, 2024.

\bibitem[\protect\citeauthoryear{Ravuri \bgroup \em et al.\egroup }{2021}]{ravuri2021skilful}
Suman Ravuri, Karel Lenc, Matthew Willson, Dmitry Kangin, Remi Lam, Piotr Mirowski, Megan Fitzsimons, Maria Athanassiadou, Sheleem Kashem, Sam Madge, et~al.
\newblock Skilful precipitation nowcasting using deep generative models of radar.
\newblock {\em Nature}, 597(7878):672--677, 2021.

\bibitem[\protect\citeauthoryear{Roberts \bgroup \em et al.\egroup }{2020}]{roberts2020much}
Adam Roberts, Colin Raffel, and Noam Shazeer.
\newblock How much knowledge can you pack into the parameters of a language model?
\newblock {\em arXiv preprint arXiv:2002.08910}, 2020.

\bibitem[\protect\citeauthoryear{Roberts \bgroup \em et al.\egroup }{2024}]{roberts2024charting}
Jonathan Roberts, Timo L{\"u}ddecke, Rehan Sheikh, Kai Han, and Samuel Albanie.
\newblock Charting new territories: Exploring the geographic and geospatial capabilities of multimodal llms.
\newblock In {\em Proceedings of the IEEE/CVF Conference on Computer Vision and Pattern Recognition}, pages 554--563, 2024.

\bibitem[\protect\citeauthoryear{Schumann \bgroup \em et al.\egroup }{2024}]{schumann2024velma}
Raphael Schumann, Wanrong Zhu, Weixi Feng, Tsu-Jui Fu, Stefan Riezler, and William~Yang Wang.
\newblock Velma: Verbalization embodiment of llm agents for vision and language navigation in street view.
\newblock In {\em Proceedings of the AAAI Conference on Artificial Intelligence}, volume~38, pages 18924--18933, 2024.

\bibitem[\protect\citeauthoryear{Shao \bgroup \em et al.\egroup }{2024}]{shao2024beyond}
Chenyang Shao, Fengli Xu, Bingbing Fan, Jingtao Ding, Yuan Yuan, Meng Wang, and Yong Li.
\newblock Beyond imitation: Generating human mobility from context-aware reasoning with large language models.
\newblock {\em arXiv preprint arXiv:2402.09836}, 2024.

\bibitem[\protect\citeauthoryear{Sharma}{2023}]{sharma2023exploring}
Manasi Sharma.
\newblock Exploring and improving the spatial reasoning abilities of large language models.
\newblock In {\em I Can't Believe It's Not Better Workshop: Failure Modes in the Age of Foundation Models}, 2023.

\bibitem[\protect\citeauthoryear{She \bgroup \em et al.\egroup }{2024}]{she2024llmdiff}
Lei She, Chenghong Zhang, Xin Man, and Jie Shao.
\newblock Llmdiff: Diffusion model using frozen llm transformers for precipitation nowcasting.
\newblock {\em Sensors}, 24(18):6049, 2024.

\bibitem[\protect\citeauthoryear{Song \bgroup \em et al.\egroup }{2024}]{song2024guide}
Sangmim Song, Sarath Kodagoda, Amal Gunatilake, Marc~G Carmichael, Karthick Thiyagarajan, and Jodi Martin.
\newblock Guide-llm: An embodied llm agent and text-based topological map for robotic guidance of people with visual impairments.
\newblock {\em arXiv preprint arXiv:2410.20666}, 2024.

\bibitem[\protect\citeauthoryear{Tolman}{1948}]{tolman1948cognitive}
Edward~C Tolman.
\newblock Cognitive maps in rats and men.
\newblock {\em Psychological review}, 55(4):189, 1948.

\bibitem[\protect\citeauthoryear{Wang \bgroup \em et al.\egroup }{2023}]{wang2023generating}
Xiaohan Wang, Yuehu Liu, Xinhang Song, Beibei Wang, and Shuqiang Jiang.
\newblock Generating explanations for embodied action decision from visual observation.
\newblock In {\em Proceedings of the 31st ACM International Conference on Multimedia}, pages 2838--2846, 2023.

\bibitem[\protect\citeauthoryear{Wang \bgroup \em et al.\egroup }{2024a}]{wang2024large}
Jiawei Wang, Renhe Jiang, Chuang Yang, Zengqing Wu, Makoto Onizuka, Ryosuke Shibasaki, Noboru Koshizuka, and Chuan Xiao.
\newblock Large language models as urban residents: An llm agent framework for personal mobility generation.
\newblock {\em arXiv preprint arXiv:2402.14744}, 2024.

\bibitem[\protect\citeauthoryear{Wang \bgroup \em et al.\egroup }{2024b}]{wang2024embodiedscan}
Tai Wang, Xiaohan Mao, Chenming Zhu, Runsen Xu, Ruiyuan Lyu, Peisen Li, Xiao Chen, Wenwei Zhang, Kai Chen, Tianfan Xue, et~al.
\newblock Embodiedscan: A holistic multi-modal 3d perception suite towards embodied ai.
\newblock In {\em Proceedings of the IEEE/CVF Conference on Computer Vision and Pattern Recognition}, pages 19757--19767, 2024.

\bibitem[\protect\citeauthoryear{Whittington \bgroup \em et al.\egroup }{2020}]{whittington2020tolman}
James~CR Whittington, Timothy~H Muller, Shirley Mark, Guifen Chen, Caswell Barry, Neil Burgess, and Timothy~EJ Behrens.
\newblock The tolman-eichenbaum machine: unifying space and relational memory through generalization in the hippocampal formation.
\newblock {\em Cell}, 183(5):1249--1263, 2020.

\bibitem[\protect\citeauthoryear{Whittington \bgroup \em et al.\egroup }{2021}]{whittington2021relating}
James~CR Whittington, Joseph Warren, and Timothy~EJ Behrens.
\newblock Relating transformers to models and neural representations of the hippocampal formation.
\newblock {\em arXiv preprint arXiv:2112.04035}, 2021.

\bibitem[\protect\citeauthoryear{Wu \bgroup \em et al.\egroup }{2024}]{wu2024torchspatial}
Nemin Wu, Qian Cao, et~al.
\newblock {T}orchspatial: {A} location encoding framework and benchmark for spatial representation learning.
\newblock In {\em Proc. of NeurIPS}, 2024.

\bibitem[\protect\citeauthoryear{Xiao \bgroup \em et al.\egroup }{2024}]{xiao2024refound}
Congxi Xiao, Jingbo Zhou, Yixiong Xiao, Jizhou Huang, and Hui Xiong.
\newblock Refound: Crafting a foundation model for urban region understanding upon language and visual foundations.
\newblock In {\em Proceedings of the 30th ACM SIGKDD Conference on Knowledge Discovery and Data Mining}, pages 3527--3538, 2024.

\bibitem[\protect\citeauthoryear{Xu \bgroup \em et al.\egroup }{2024a}]{xu2024flame}
Yunzhe Xu, Yiyuan Pan, Zhe Liu, and Hesheng Wang.
\newblock Flame: Learning to navigate with multimodal llm in urban environments.
\newblock {\em arXiv preprint arXiv:2408.11051}, 2024.

\bibitem[\protect\citeauthoryear{Xu \bgroup \em et al.\egroup }{2024b}]{xu2024geopredict}
Zhenhao Xu, Zhaoyang Wang, Shucai Li, Xiao Zhang, and Peng Lin.
\newblock Geopredict-llm: Intelligent tunnel advanced geological prediction by reprogramming large language models.
\newblock {\em Intelligent Geoengineering}, 1(1):49--57, 2024.

\bibitem[\protect\citeauthoryear{Xu \bgroup \em et al.\egroup }{2025}]{xu2025defining}
Wenrui Xu, Dalin Lyu, Weihang Wang, Jie Feng, Chen Gao, and Yong Li.
\newblock Defining and evaluating visual language models' basic spatial abilities: A perspective from psychometrics.
\newblock {\em arXiv preprint arXiv:2502.11859}, 2025.

\bibitem[\protect\citeauthoryear{Yamada \bgroup \em et al.\egroup }{2023}]{yamada2023evaluating}
Yutaro Yamada, Yihan Bao, Andrew~K Lampinen, Jungo Kasai, and Ilker Yildirim.
\newblock Evaluating spatial understanding of large language models.
\newblock {\em arXiv preprint arXiv:2310.14540}, 2023.

\bibitem[\protect\citeauthoryear{Yan and Lee}{2024}]{yan2024georeasoner}
Yibo Yan and Joey Lee.
\newblock Georeasoner: Reasoning on geospatially grounded context for natural language understanding.
\newblock In {\em Proceedings of the 33rd ACM International Conference on Information and Knowledge Management}, pages 4163--4167, 2024.

\bibitem[\protect\citeauthoryear{Yan \bgroup \em et al.\egroup }{2024}]{yan2024urbanclip}
Yibo Yan, Haomin Wen, Siru Zhong, Wei Chen, Haodong Chen, Qingsong Wen, Roger Zimmermann, and Yuxuan Liang.
\newblock Urbanclip: Learning text-enhanced urban region profiling with contrastive language-image pretraining from the web.
\newblock In {\em Proceedings of the ACM on Web Conference 2024}, pages 4006--4017, 2024.

\bibitem[\protect\citeauthoryear{Yang \bgroup \em et al.\egroup }{2024a}]{yang2024llmi3d}
Fan Yang, Sicheng Zhao, Yanhao Zhang, Haoxiang Chen, Hui Chen, Wenbo Tang, Haonan Lu, Pengfei Xu, Zhenyu Yang, Jungong Han, et~al.
\newblock Llmi3d: Empowering llm with 3d perception from a single 2d image.
\newblock {\em arXiv preprint arXiv:2408.07422}, 2024.

\bibitem[\protect\citeauthoryear{Yang \bgroup \em et al.\egroup }{2024b}]{yang2024thinking}
Jihan Yang, Shusheng Yang, Anjali~W. Gupta, Rilyn Han, Li~Fei-Fei, and Saining Xie.
\newblock Thinking in space: How multimodal large language models see, remember, and recall spaces, 2024.

\bibitem[\protect\citeauthoryear{Yang \bgroup \em et al.\egroup }{2024c}]{yang2024oceanplan}
Ruochu Yang, Fumin Zhang, and Mengxue Hou.
\newblock Oceanplan: Hierarchical planning and replanning for natural language auv piloting in large-scale unexplored ocean environments.
\newblock {\em arXiv preprint arXiv:2403.15369}, 2024.

\bibitem[\protect\citeauthoryear{Yang \bgroup \em et al.\egroup }{2024d}]{yang20243dmem}
Yuncong Yang, Han Yang, Jiachen Zhou, Peihao Chen, Hongxin Zhang, Yilun Du, and Chuang Gan.
\newblock 3d-mem: 3d scene memory for embodied exploration and reasoning.
\newblock {\em arXiv preprint arXiv:2411.17735}, 2024.

\bibitem[\protect\citeauthoryear{Yu \bgroup \em et al.\egroup }{2024a}]{yu2024mineagent}
Beibei Yu, Tao Shen, Hongbin Na, Ling Chen, and Denqi Li.
\newblock Mineagent: Towards remote-sensing mineral exploration with multimodal large language models.
\newblock {\em arXiv preprint arXiv:2412.17339}, 2024.

\bibitem[\protect\citeauthoryear{Yu \bgroup \em et al.\egroup }{2024b}]{yu2024rag}
Jun Yu, Yunxiang Zhang, Zerui Zhang, Zhao Yang, Gongpeng Zhao, Fengzhao Sun, Fanrui Zhang, Qingsong Liu, Jianqing Sun, Jiaen Liang, et~al.
\newblock Rag-guided large language models for visual spatial description with adaptive hallucination corrector.
\newblock In {\em Proceedings of the 32nd ACM International Conference on Multimedia}, pages 11407--11413, 2024.

\bibitem[\protect\citeauthoryear{Zeng \bgroup \em et al.\egroup }{2024}]{zeng2024perceive}
Qingbin Zeng, Qinglong Yang, Shunan Dong, Heming Du, Liang Zheng, Fengli Xu, and Yong Li.
\newblock Perceive, reflect, and plan: Designing llm agent for goal-directed city navigation without instructions.
\newblock {\em arXiv preprint arXiv:2408.04168}, 2024.

\bibitem[\protect\citeauthoryear{Zhang \bgroup \em et al.\egroup }{2023a}]{zhang2023geogpt}
Yifan Zhang, Cheng Wei, Shangyou Wu, Zhengting He, and Wenhao Yu.
\newblock Geogpt: understanding and processing geospatial tasks through an autonomous gpt.
\newblock {\em arXiv preprint arXiv:2307.07930}, 2023.

\bibitem[\protect\citeauthoryear{Zhang \bgroup \em et al.\egroup }{2023b}]{zhang2023skilful}
Yuchen Zhang, Mingsheng Long, Kaiyuan Chen, Lanxiang Xing, Ronghua Jin, Michael~I Jordan, and Jianmin Wang.
\newblock Skilful nowcasting of extreme precipitation with nowcastnet.
\newblock {\em Nature}, 619(7970):526--532, 2023.

\bibitem[\protect\citeauthoryear{Zhang \bgroup \em et al.\egroup }{2023c}]{zhang2023large}
Zihan Zhang, Meng Fang, Ling Chen, Mohammad-Reza Namazi-Rad, and Jun Wang.
\newblock How do large language models capture the ever-changing world knowledge? a review of recent advances.
\newblock {\em arXiv preprint arXiv:2310.07343}, 2023.

\bibitem[\protect\citeauthoryear{Zhang \bgroup \em et al.\egroup }{2024}]{zhang2024geoeval}
Jiaxin Zhang, Zhongzhi Li, Mingliang Zhang, Fei Yin, Chenglin Liu, and Yashar Moshfeghi.
\newblock Geoeval: benchmark for evaluating llms and multi-modal models on geometry problem-solving.
\newblock {\em arXiv preprint arXiv:2402.10104}, 2024.

\bibitem[\protect\citeauthoryear{Zhao \bgroup \em et al.\egroup }{2024}]{zhao2024artificial}
Tianjie Zhao, Sheng Wang, et~al.
\newblock {A}rtificial intelligence for geoscience: {P}rogress, challenges and perspectives.
\newblock {\em The Innovation}, 2024.

\bibitem[\protect\citeauthoryear{Zheng \bgroup \em et al.\egroup }{2024}]{zheng2024video}
Duo Zheng, Shijia Huang, and Liwei Wang.
\newblock Video-3d llm: Learning position-aware video representation for 3d scene understanding.
\newblock {\em arXiv preprint arXiv:2412.00493}, 2024.

\bibitem[\protect\citeauthoryear{Zhong \bgroup \em et al.\egroup }{2024}]{zhong2024topv}
Linqing Zhong, Chen Gao, Zihan Ding, Yue Liao, and Si~Liu.
\newblock Topv-nav: Unlocking the top-view spatial reasoning potential of mllm for zero-shot object navigation.
\newblock {\em arXiv preprint arXiv:2411.16425}, 2024.

\bibitem[\protect\citeauthoryear{Zhou \bgroup \em et al.\egroup }{2024a}]{zhou2024navgpt}
Gengze Zhou, Yicong Hong, and Qi~Wu.
\newblock Navgpt: Explicit reasoning in vision-and-language navigation with large language models.
\newblock In {\em Proceedings of the AAAI Conference on Artificial Intelligence}, volume~38, pages 7641--7649, 2024.

\bibitem[\protect\citeauthoryear{Zhou \bgroup \em et al.\egroup }{2024b}]{zhou2024large}
Zhilun Zhou, Yuming Lin, Depeng Jin, and Yong Li.
\newblock Large language model for participatory urban planning.
\newblock {\em arXiv preprint arXiv:2402.17161}, 2024.

\bibitem[\protect\citeauthoryear{Zhou \bgroup \em et al.\egroup }{2025}]{zhou2025navgpt}
Gengze Zhou, Yicong Hong, Zun Wang, Xin~Eric Wang, and Qi~Wu.
\newblock Navgpt-2: Unleashing navigational reasoning capability for large vision-language models.
\newblock In {\em European Conference on Computer Vision}, pages 260--278. Springer, 2025.

\bibitem[\protect\citeauthoryear{Zitkovich \bgroup \em et al.\egroup }{2023}]{zitkovich2023rt}
Brianna Zitkovich, Tianhe Yu, Sichun Xu, Peng Xu, Ted Xiao, Fei Xia, Jialin Wu, Paul Wohlhart, Stefan Welker, Ayzaan Wahid, et~al.
\newblock Rt-2: Vision-language-action models transfer web knowledge to robotic control.
\newblock In {\em Conference on Robot Learning}, pages 2165--2183. PMLR, 2023.

\end{thebibliography}

\end{document}